\documentclass[10pt,twocolumn,letterpaper]{article}

\usepackage{cvpr}
\usepackage{times}
\usepackage{epsfig}
\usepackage{graphicx}
\usepackage{amsmath}
\usepackage{amssymb}

\usepackage[breaklinks=true,bookmarks=false]{hyperref}

\cvprfinalcopy 

\def\cvprPaperID{****} 

\begin{document}

\title{End-to-end Convolutional Network for Saliency Prediction}

\author{Junting Pan and Xavier Gir\'o-i-Nieto\\
Universitat Politecnica de Catalunya (UPC)\\
Barcelona, Catalonia/Spain\\
{\tt\small junting.pan@alu-etsetb.upc.edu, xavier.giro@upc.edu}
}

\maketitle

\begin{abstract}
The prediction of saliency areas in images has been traditionally addressed with hand crafted features based on neuroscience principles.
This paper however addreses the problem with a completely data-driven approach by training a convolutional network.
The learning process is formulated as a minimization of a loss function that measures the Euclidean distance of the predicted saliency map with the provided ground truth.
The recent publication of large datasets of saliency prediction has provided enough data to train a not very deep architecture which is both fast and accurate.
The convolutional network in this paper, named JuntingNet, won the LSUN 2015 challenge on saliency prediction with a superior performance in all considered metrics.
\end{abstract}
\section{Introduction}

This work presents an end-to-end convolutional network (convnet) for saliency prediction. 
Our objective is to compute saliency maps that represent the probability of visual attention. 
This problem has been traditionally addressed with hand-crafted features inspired by neurology studies.
In our case we have adopted a completely data-driven approach, training a model with a large amount of annotated data.

Convnet is a popular architecture in the field of deep learning and has been widely explored for visual pattern recognition, ranging from a global scale image classification to a more local object detection or semantic segmentation.
The hierarchy of layers of convnets are also inspired by biological models and actually recent works have pointed at a relation between the activity of certain areas in the brain with hierarchy of layers in the convnets \cite{agrawal2014pixels}.
Provided with enough training data, convnets show impressive results, often outperforming other hand-crafted methods . 
In many popular works, the output of the convnet is a discrete label associated to a certain semantic class.
The saliency prediction problem, though, addresses the problem of a continuous range of values that estimate the probability of a human fixation on a pixel.
These values present a spatial coherence and smooth transition that this work addresses by using the convnet as a regression solver, instead of a classifier. 

\begin{figure}%
		\includegraphics[width=\linewidth]{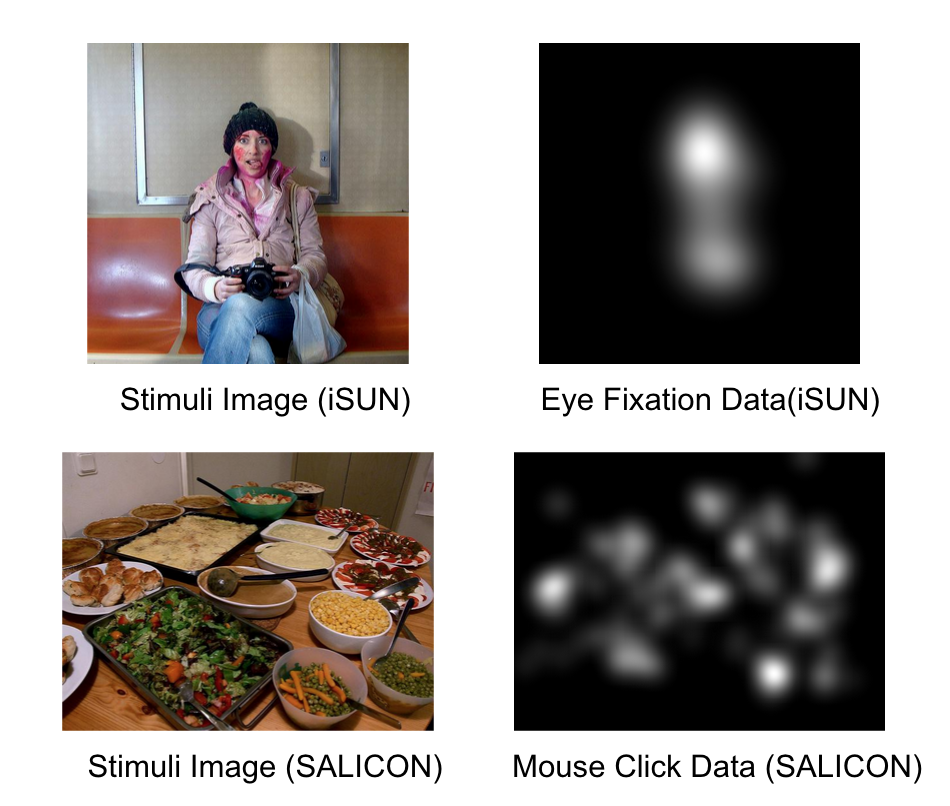}
		\caption{Images (right) and saliency maps (left) from the iSUN and SALICON datasets.}
		\label{fig:examples}
\end{figure}

The training of a convolutional network requires a large amount of annotated data that provides a rich description of the problem.
Our work has benefited from the recent publication of two datasets: iSun \cite{xu2015turkergaze} and SALICON \cite{jiang2015salicon}.
These datasets propose two different approaches for saliency prediction.
While iSun was generated with an eye-tracker to annotate the gaze fixations, the SALICON dataset was built by asking humans to click on the most salient points on the image.
The different nature of the saliency maps of the two datasets can be seen in Figure \ref{fig:examples}.
The large size of these datasets has provided for the first time the possibility of training a convnet.

Our main contribution has been the design of an end-to-end convnet for saliency prediction, the first one from this type, up to the authors knowledge.
The network, called \textit{JuntingNet}, has proved its superior performance in the Large-scale Scene UNderstanding (LSUN) challenge 2015 \cite{zhanglarge}.
The developed model has been publicly available at \url{http://bit.ly/juntingnet}.




This paper is structured as follows. Section \ref{sec:RelatedWork} presents the previous works using convolutional networks for saliency prediction.
Our system is presented in Section \ref{sec:juntingnet} and its results on the LSUN challenge reported in Section \ref{sec:experiments}.
The conclusions and future directions are contained in Section \ref{sec:conclusions}.
\section{Related work}
\label{sec:RelatedWork}

JuntingNet presents the next natural step to two main trends in deep learning: using convnets for saliency prediction and training these networks by formulating and end-to-end problem.
This section refers to some related work in these two fields.


\subsection{Deep learning for saliency prediction}

An early attempt of predicting saliency model with a convnet was the \textit{ensembles of Deep Networks (eDN)} \cite{vig2014large}, which proposed an optimal blend of features from three different convnet layers who were finally combined with a simple linear classifier trained with positive (salient) or negative (non-salient) local regions.
This approach inspired \textit{DeepGaze} \cite{kummerer2014deep}, which only combined features from different layers but, in this case, from a much deeper network.
In particular, \textit{DeepGaze} used the existing \textit{AlexNet} convnet \cite{krizhevsky2012imagenet}, which had been trained for an object classification task, not for saliency prediction. 
\textit{JuntingNet} adopts a not very deep architecture as \textit{eDN}, but it is end-to-end trained as a regression problem, avoiding the reuse of precomputed parameters from another task.

\subsection{End to end semantic segmentation}

Fully Convolutional Networks (FCNs) \cite{DBLP:journals/corr/LongSD14} addressed the  semantic segmentation task which predicting the semantic label of every individual pixel in the image. 
This approach dramatically improved previous results on the challenging PASCAL VOC segmentation benchmark \cite{everingham2014pascal} .
The idea of an end-to-end solution for a 2D problem as as semantic segmentation was refined by \text{DeepLab-CRF} \cite{DBLP:journals/corr/ChenPKMY14}, where the spatial consistency of the predicted labels is checked with a Conditional Random Field (CRF), similarly to the hierarchical consistency enforced in \cite{farabet2013learning}.
In our work, we adopt the end-to-end solution for a regression problem instead of a classification one, and we also introduce a post-filtering stage, which consists of a Gaussian filtering that smoothes the resulting saliency map.

\section{JuntingNet}
\label{sec:juntingnet}

This paper presents \textit{JuntingNet}, an end-to-end convnet for saliency prediction.
The parameters of our network are learned by minimizing an Euclidean loss function defined directly on the ground truth saliency maps.


\subsection{Architecture}

\begin{figure*}
%
%
%
		\includegraphics[width=\textwidth]{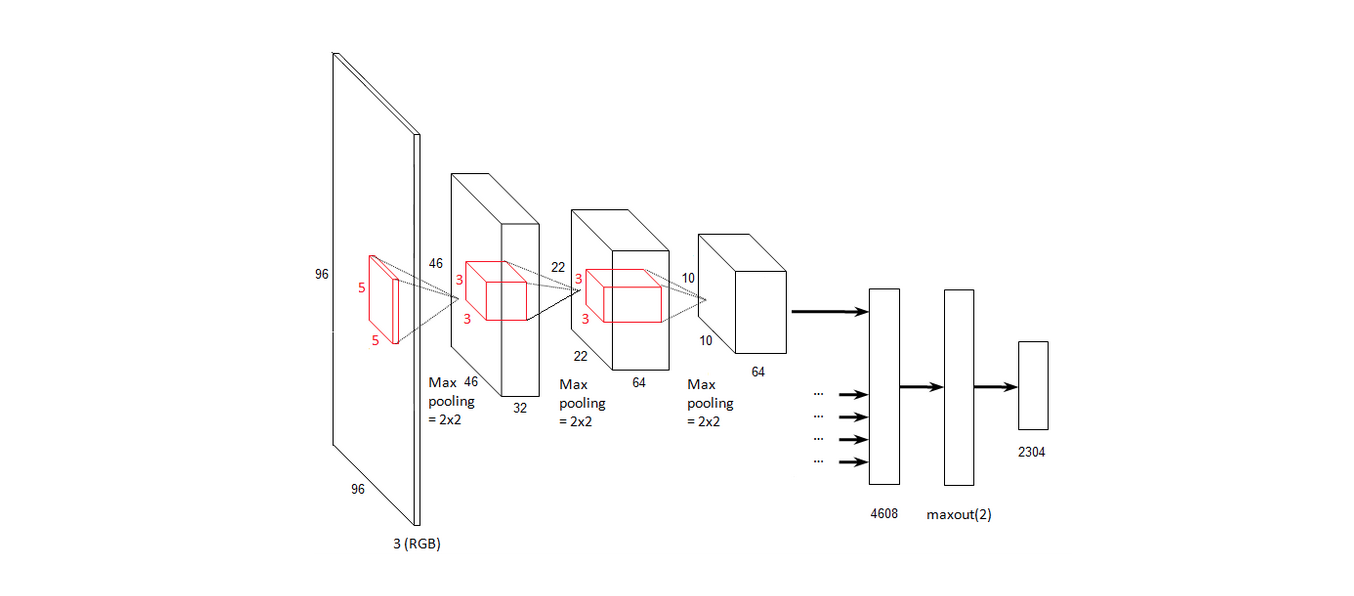}
		\caption{Convnet architecture for JuntingNet.}
		\label{fig:juntingnet}
\end{figure*}

The detailed architecture of JuntingNet is illustrated in Figure \ref{fig:juntingnet}. 
The network contains five learned layers: three convolutional layers and two fully connected layers, which can also be interpreted as 1x1 convolutions.

The proposed architecture is not very deep if compared to other networks in the state of the art.
Popular architectures trained on the $1,200,000$ images of the ILSRVC 2012 challenge proposed from 7 \cite{krizhevsky2012imagenet} to 22 layers \cite{szegedy2014going}.
\textit{JuntingNet} is defined by only 5 layers which are trained separately on two training datasets collections of diverse sizes: $6,000$ for iSun and $10,000$ for SALICON.
This adopted shallow depth tries to prevent the overfitting problem, which is a great risk for models with a large amount of parameters, such as convnets.


The detailed description of the convnet stages is the following:

\begin{enumerate}

\item The input volume has size of [96x96x3] (RGB image), a size smaller than the [227x227x3] proposed in AlexNet. Similarly to the shallow depth, this design parameter is motivated to reduce the possibilities of overfitting. 

\item The receptive field of the first 2D convolution is of size [5x5], and its outputs define a convolutional layer with 32 neurons.
This layer is followed by a ReLU activation layer which applies an element wise non-linearity.
Later, a max pooling layer progressively reduces the spatial size of the input image. 
Despite the loss of visual resolution at the output, this reduction also reduces the amount of model parameters and  prevents overfitting. 
The max-pooling layer selects the maximum value of every [2x2] region, taking strides of two pixels. 

\item The output of the previous stage has a size of [46x46x32]. 
The receptive field of this second stage is [3x3]. Again, this is followed by a RELU layer and a max-pooling layer of size [2x2].

\item Finally, the last convolutional layer is fed with an input of size [22x22x64]. 
The receptive of this layer is also of [3x3] and it has 64 neurons. 
A ReLU and max pooling layers are stacked too. 

\item A first fully connected layer receives the output of the third convolutional layer with a dimension of [10x10x64]. It contains a total of 4,608 neurons.

\item The second fully connected layer consist of a maxout layer with 2,304 neurons.
The maxout operation \cite{goodfellow2013maxout} computes the pairs of the previous layer’s output. 

\item Finally, the output of the last maxout layer is the saliency prediction array. 
The array is reshaped to have 2D dimensions and resized to the stimuli image size. 
Finally, a 2D Gaussian filter with a standard deviation of 3.0 is applied.

\end{enumerate}

\subsection{Training parameters}

The limited amount of training data for our architecture made overfitting a significant challenge, so we used different techniques to minimize its effects. 
Firstly, we apply norm constraint regularization for the maxout layers \cite{goodfellow2013maxout}. 
Secondly, we use data augmentation technique by mirroring all images. 
We also tested a dropout layer \cite{hinton2012improving} after the first fully connected layer, with a dropout ratio of 0.5 (50\% of probability to set a neuron’s output value to zero). 
However, this did not make much of a difference, so it is not included to the final model.

The weights in all layers are initialized from a normal Gaussian distribution with zero mean and a standard deviation of 0.01, with biases initialized to 0.1.
Ground truth values that we used for training are saliency maps with normalized values between 0 and 1.

\begin{figure}%
		\includegraphics[width=\linewidth]{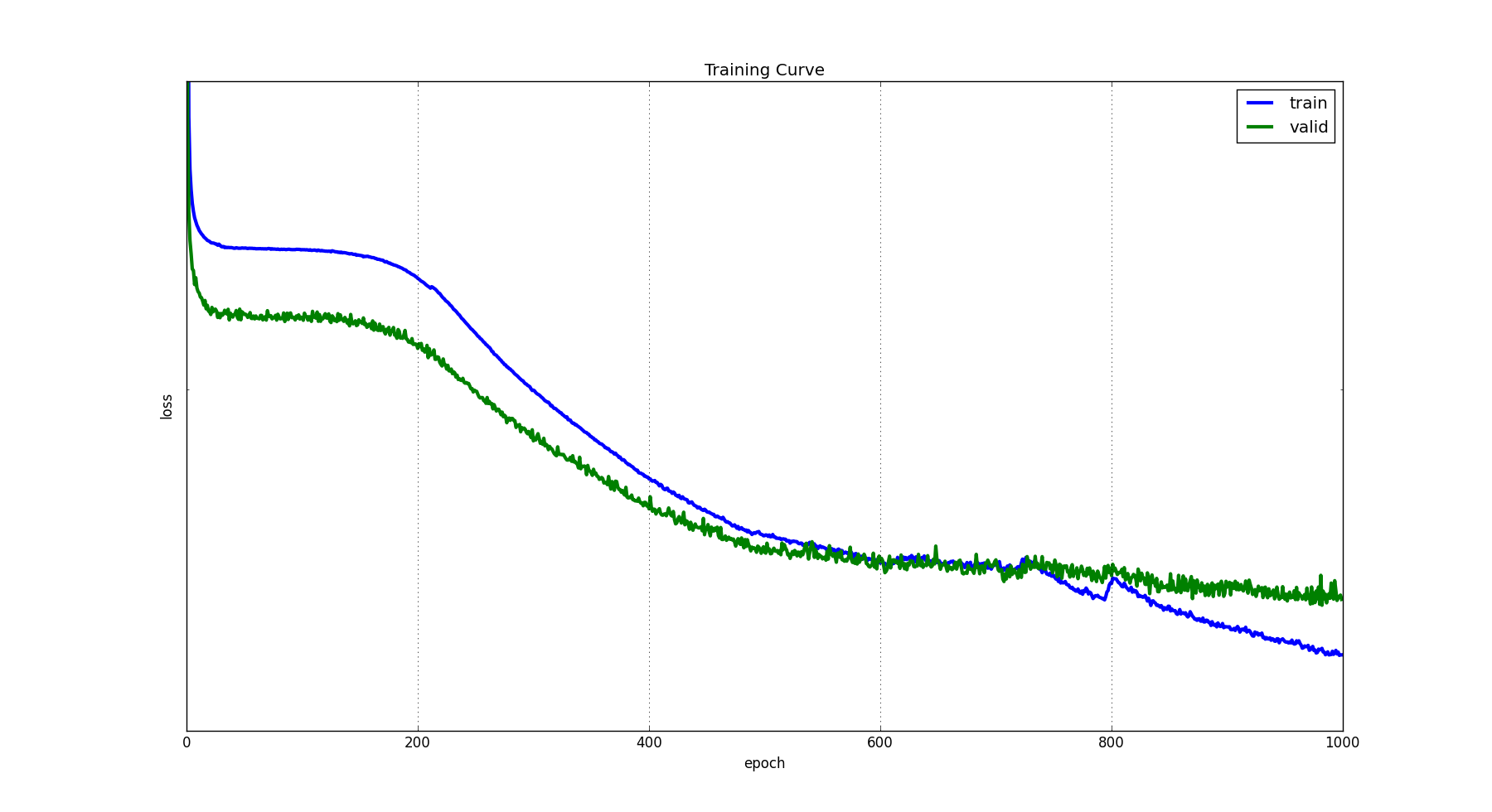}
		\caption{Learning curves for iSUN models.}
		\label{fig:train-isun}
\end{figure}

\begin{figure}%
		\includegraphics[width=\linewidth]{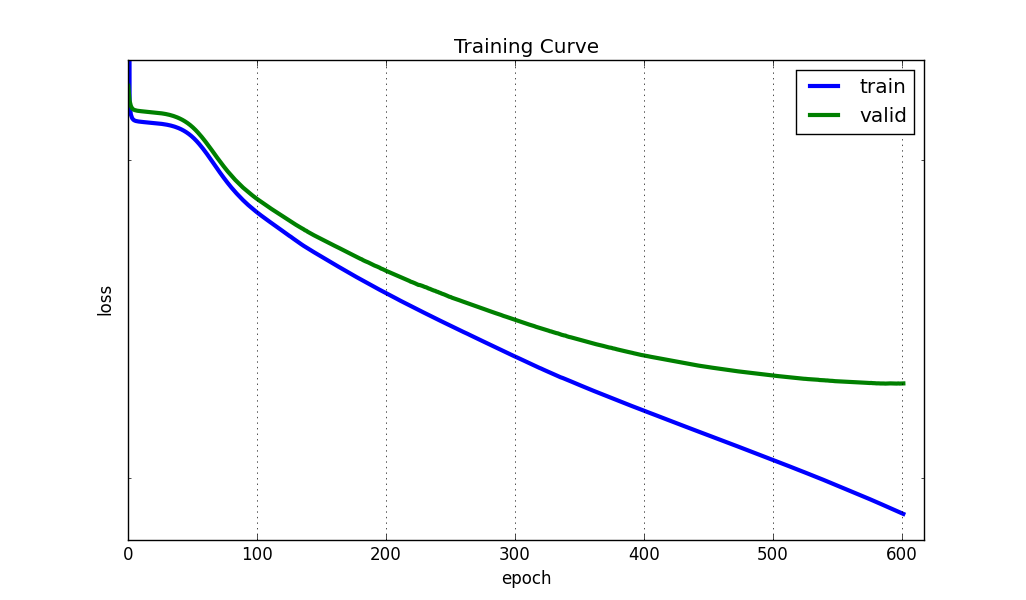}
		\caption{Learning curves for SALICON models.}
		\label{fig:train-salicon}
\end{figure}

For validation control purposes, we split the training partitions of iSUN and SALICON datasets into 80\% for training and the rest for real time validation. 
The network was trained with stochastic gradient descent (SGD) and Nesterov momentum SGD optimization method that helps the loss function to converge faster. 
The learning rate was changing over time; it started with a higher learning rate 0.03 and decreased during the course of training until 0.0001. 
We set 1,000 epochs to train a separate network for each dataset.
Figures \ref{fig:train-isun} and \ref{fig:train-salicon} present the learning curves for the iSUN and SALICON models, respectively.

\section{Experiments}
\label{sec:experiments}

\subsection{Datasets}

The network was tested in the two datasets proposed in the LSUN challenge \cite{zhanglarge}: 

\begin{description}

\item [iSUN \cite{xu2015turkergaze}:] a ground truth of gaze traces on images from the SUN dataset \cite{xiao2010sun}. 
The collection is partitioned into 6,000 images for training, 926 for validation and 2,000 for test.

\item [SALICON \cite{jiang2015salicon}:] cursor clicks on the objects of interest from images of the Microsoft COCO dataset \cite{lin2014microsoft}. 
The collection contains 10,000 training images, 5,000 for validation and 5,000 for test.
\end{description}

\subsection{Results}

Our solution is implemented using Python, NumPy and the deep learning library Theano \cite{bergstra2010theano, bastien2012theano}. 
Processing was performed on an NVidia GPU GTX 980 with 2048 CUDA cores and 4GB of RAM. 
Our network took between six to seven hours to train for the SALICON dataset, and five to six hours for the iSUN dataset.
Every saliency prediction requires 200 ms per image.

We assessed our model on the LSUN saliency prediction challenge 2015 \cite{zhanglarge}. 
Table \ref{tab:iSUN} and Table \ref{tab:SALICON} presents our results for iSUN and SALICON datasets. 
The model was evaluated separately on the testing data of each datasets. 
The evaluation metrics was adopted of the variety of metrics provided in MIT saliency benchmark \cite{Judd_2012, mit-saliency-benchmark} defined on both saliency map and fixation points. 
\textit{JuntingNet} consistenly won the first place of the challenge in all metrics considered in the challenge. A few qualitative results are also provided in Figure \ref{fig:juntingnet}.

\begin{table*}
\begin{center}
\begin{tabular}{|l|c|c|c|c|c|}
\hline
					&	Similarity 	& CC 		& AUC shuffled 	& AUC Borji & AUC Judd \\
\hline\hline
\textbf{Our work}	& \textbf{$0.6833$} 		& \textbf{$0.8230$}	& \textbf{$0.6650$} 		& \textbf{$0.8463$}	& \textbf{$0.8693$} \\
Xidian 				& $0.5713$ 		& $0.6167$	& $0.6484$ 		& $0.7949$	& $0.8207$ \\
WHU IIP 			& $0.5593$ 		& $0.6263$	& $0.6307$ 		& $0.7960$	& $0.8197$ \\
LCYLab 				& $0.5474$ 		& $0.5699$	& $0.6259$ 		& $0.7921$	& $0.8133$ \\
Rare 2012 Improved 	& $0.5199$ 		& $0.5199$	& $0.6283$ 		& $0.7582$	& $0.7846$ \\
\hline
Baseline: BMS \cite{zhang2013saliency}		& $0.5026$ 		& $0.3465$	& $0.5885$ 		& $0.6560$	& $0.6914$ \\
Baseline: GBVS \cite{harel2006graph}		& $0.4798$ 		& $0.5087$	& $0.6208$ 		& $0.7913$	& $0.8115$ \\
Baseline: Itti \cite{itti1998model}		& $0.4251$ 		& $0.3728$	& $0.6024$ 		& $0.7262$	& $0.7489$ \\
\hline
\end{tabular}
\end{center}
\caption{Results of the LSUN challenge 2015 for saliency prediction with the iSUN dataset.}
\label{tab:iSUN}
\end{table*}

\begin{table*}
\begin{center}
\begin{tabular}{|l|c|c|c|c|c|}
\hline
					&	Similarity 	& CC 		& AUC shuffled 	& AUC Borji & AUC Judd \\
\hline\hline
\textbf{Our work}	& \textbf{$0.5198$} 		& \textbf{$0.5957$}	& \textbf{$0.6698$} 		& \textbf{$0.8291$}	& \textbf{$0.8364$} \\
WHU IIP 			& $0.4908$ 		& $0.4569$	& $0.6064$ 		& $0.7759$	& $0.7923$ \\
Rare 2012 Improved 	& $0.5017$ 		& $0.5108$	& $0.6644$ 		& $0.8047$	& $0.8148$ \\
Xidian			 	& $0.4617$ 		& $0.4811$	& $0.6809$ 		& $0.7990$	& $0.8051$ \\
\hline
Baseline: BMS \cite{zhang2013saliency}		& $0.4542$ 		& $0.4268$	& $0.6935$ 		& $0.7699$	& $0.7899$ \\
Baseline: GBVS \cite{harel2006graph}		& $0.4460$ 		& $0.4212$	& $0.6303$ 		& $0.7816$	& $0.7899$ \\
Baseline: Itti \cite{itti1998model}		& $0.3777$ 		& $0.2046$	& $0.6101$ 		& $0.6603$	& $0.6669$ \\
\hline
\end{tabular}
\end{center}
\caption{Results of the LSUN challenge 2015 for saliency prediction with the SALICON dataset.}
\label{tab:SALICON}
\end{table*}

\begin{figure*}
		\includegraphics[width=\textwidth]{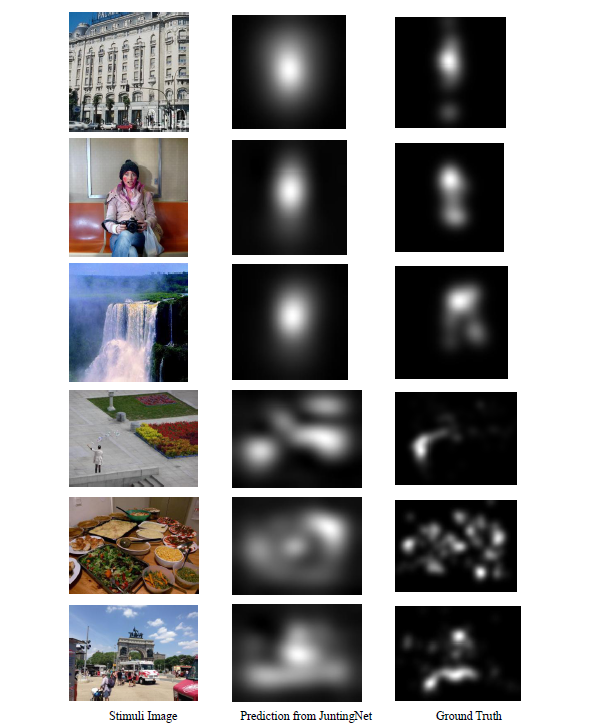}
		\caption{Saliency maps generated by JuntingNet. The first column corresponds to the input image, the second column the prediction from JuntingNet and the third one and the third on to the provided ground truth. First three rows correspond to images from the iSUN dataset, while the last three are from the SALICON dataset.}
		\label{fig:qualitative}
\end{figure*}

\section{Conclusions}
\label{sec:conclusions}

We designed the first end-to-end ConvNet for saliency prediction, trained only with the datasets of visual saliency provided by the LSUN challenge. 
With this ConvNet we were able to win the first place in the challenge by large margin. 
Our results demonstrate that a not very deep ConvNets are capable of achieving good results on a highly challenging task.  

Our experiments can be considered as preliminary, as only one configuration and set up was considered.
We expect that a more elaborate study of the architecture, use of the dataset and training parameters could still improve the reported performance.





The developed model has been publicly availble from \url{http://bit.ly/juntingnet}.
\section{Acnowledgements}

We would like to thank the technical support of Albert Gil and Josep Pujal in the setting up of software and hardware necessary to run the experimentation.

The Image Processing Group at the UPC is a SGR14 Consolidated Research Group recognized and sponsored by the Catalan Government (Generalitat de Catalunya) through its  AGAUR office.

This work has been developed in the framework of the project BigGraph TEC2013-43935-R, funded by the Spanish Ministerio de Economía y Competitividad and the European Regional Development Fund (ERDF). 

We gratefully acknowledge the support of NVIDIA Corporation with the donation of the GeoForce GTX 980 used in this work.

{\small
\bibliographystyle{ieee}
\bibliography{egbib}

\begin{thebibliography}{10}\itemsep=-1pt

\bibitem{agrawal2014pixels}
P.~Agrawal, D.~Stansbury, J.~Malik, and J.~L. Gallant.
\newblock Pixels to voxels: Modeling visual representation in the human brain.
\newblock {\em arXiv preprint arXiv:1407.5104}, 2014.

\bibitem{bastien2012theano}
F.~Bastien, P.~Lamblin, R.~Pascanu, J.~Bergstra, I.~Goodfellow, A.~Bergeron,
  N.~Bouchard, D.~Warde-Farley, and Y.~Bengio.
\newblock Theano: new features and speed improvements.
\newblock {\em arXiv preprint arXiv:1211.5590}, 2012.

\bibitem{bergstra2010theano}
J.~Bergstra, O.~Breuleux, F.~Bastien, P.~Lamblin, R.~Pascanu, G.~Desjardins,
  J.~Turian, D.~Warde-Farley, and Y.~Bengio.
\newblock Theano: a cpu and gpu math expression compiler.
\newblock In {\em Proceedings of the Python for scientific computing conference
  (SciPy)}, volume~4, page~3. Austin, TX, 2010.

\bibitem{mit-saliency-benchmark}
Z.~Bylinskii, T.~Judd, A.~Borji, L.~Itti, F.~Durand, A.~Oliva, and A.~Torralba.
\newblock Mit saliency benchmark.
\newblock http://saliency.mit.edu/.

\bibitem{DBLP:journals/corr/ChenPKMY14}
L.~Chen, G.~Papandreou, I.~Kokkinos, K.~Murphy, and A.~L. Yuille.
\newblock Semantic image segmentation with deep convolutional nets and fully
  connected crfs.
\newblock {\em CoRR}, abs/1412.7062, 2014.

\bibitem{everingham2014pascal}
M.~Everingham, S.~A. Eslami, L.~Van~Gool, C.~K. Williams, J.~Winn, and
  A.~Zisserman.
\newblock The pascal visual object classes challenge: A retrospective.
\newblock {\em International Journal of Computer Vision}, 111(1):98--136, 2014.

\bibitem{farabet2013learning}
C.~Farabet, C.~Couprie, L.~Najman, and Y.~LeCun.
\newblock Learning hierarchical features for scene labeling.
\newblock {\em Pattern Analysis and Machine Intelligence, IEEE Transactions
  on}, 35(8):1915--1929, 2013.

\bibitem{goodfellow2013maxout}
I.~J. Goodfellow, D.~Warde-Farley, M.~Mirza, A.~Courville, and Y.~Bengio.
\newblock Maxout networks.
\newblock {\em arXiv preprint arXiv:1302.4389}, 2013.

\bibitem{harel2006graph}
J.~Harel, C.~Koch, and P.~Perona.
\newblock Graph-based visual saliency.
\newblock In {\em Advances in neural information processing systems}, pages
  545--552, 2006.

\bibitem{hinton2012improving}
G.~E. Hinton, N.~Srivastava, A.~Krizhevsky, I.~Sutskever, and R.~R.
  Salakhutdinov.
\newblock Improving neural networks by preventing co-adaptation of feature
  detectors.
\newblock {\em arXiv preprint arXiv:1207.0580}, 2012.

\bibitem{itti1998model}
L.~Itti, C.~Koch, and E.~Niebur.
\newblock A model of saliency-based visual attention for rapid scene analysis.
\newblock {\em IEEE Transactions on Pattern Analysis \& Machine Intelligence},
  (11):1254--1259, 1998.

\bibitem{jiang2015salicon}
M.~Jiang, S.~Huang, J.~Duan, and Q.~Zhao.
\newblock {SALICON}: Saliency in context.
\newblock In {\em Proceedings of the {IEEE} Conf. on Computer Vision and
  Pattern Recognition ({CVPR})}, 2015.

\bibitem{Judd_2012}
T.~Judd, F.~Durand, and A.~Torralba.
\newblock A benchmark of computational models of saliency to predict human
  fixations.
\newblock In {\em MIT Technical Report}, 2012.

\bibitem{krizhevsky2012imagenet}
A.~Krizhevsky, I.~Sutskever, and G.~E. Hinton.
\newblock Imagenet classification with deep convolutional neural networks.
\newblock In {\em Advances in neural information processing systems}, pages
  1097--1105, 2012.

\bibitem{kummerer2014deep}
M.~K{\"u}mmerer, L.~Theis, and M.~Bethge.
\newblock Deep gaze i: Boosting saliency prediction with feature maps trained
  on imagenet.
\newblock {\em arXiv preprint arXiv:1411.1045}, 2014.

\bibitem{lin2014microsoft}
T.-Y. Lin, M.~Maire, S.~Belongie, J.~Hays, P.~Perona, D.~Ramanan,
  P.~Doll{\'a}r, and C.~L. Zitnick.
\newblock Microsoft coco: Common objects in context.
\newblock In {\em Computer Vision--ECCV 2014}, pages 740--755. Springer, 2014.

\bibitem{DBLP:journals/corr/LongSD14}
J.~Long, E.~Shelhamer, and T.~Darrell.
\newblock Fully convolutional networks for semantic segmentation.
\newblock {\em CoRR}, abs/1411.4038, 2014.

\bibitem{szegedy2014going}
C.~Szegedy, W.~Liu, Y.~Jia, P.~Sermanet, S.~Reed, D.~Anguelov, D.~Erhan,
  V.~Vanhoucke, and A.~Rabinovich.
\newblock Going deeper with convolutions.
\newblock {\em arXiv preprint arXiv:1409.4842}, 2014.

\bibitem{vig2014large}
E.~Vig, M.~Dorr, and D.~Cox.
\newblock Large-scale optimization of hierarchical features for saliency
  prediction in natural images.
\newblock In {\em Computer Vision and Pattern Recognition (CVPR), 2014 IEEE
  Conference on}, pages 2798--2805. IEEE, 2014.

\bibitem{xiao2010sun}
J.~Xiao, J.~Hays, K.~Ehinger, A.~Oliva, A.~Torralba, et~al.
\newblock Sun database: Large-scale scene recognition from abbey to zoo.
\newblock In {\em Computer vision and pattern recognition (CVPR), 2010 IEEE
  conference on}, pages 3485--3492. IEEE, 2010.

\bibitem{xu2015turkergaze}
P.~Xu, K.~A. Ehinger, Y.~Zhang, A.~Finkelstein, S.~R. Kulkarni, and J.~Xiao.
\newblock Turkergaze: Crowdsourcing saliency with webcam based eye tracking.
\newblock {\em arXiv preprint arXiv:1504.06755}, 2015.

\bibitem{zhang2013saliency}
J.~Zhang and S.~Sclaroff.
\newblock Saliency detection: A boolean map approach.
\newblock In {\em Computer Vision (ICCV), 2013 IEEE International Conference
  on}, pages 153--160. IEEE, 2013.

\bibitem{zhanglarge}
Y.~Zhang, F.~Yu, S.~Song, P.~Xu, A.~Seff, and J.~Xiao.
\newblock Large-scale scene understanding challenge: Eye tracking saliency
  estimation.

\end{thebibliography}
}

\end{document}